\documentclass[runningheads]{llncs}
\usepackage[T1]{fontenc}
\usepackage{graphicx}
\usepackage{amsmath}

\newcommand{\bhline}[1]{\noalign{\hrule height #1}}

\begin{document}
%
\title{ZEAL: Surgical Skill Assessment with \\ Zero-shot Surgical Tool Segmentation Using \\ Unified Foundation Model}
\titlerunning{ZEAL: Surgical Skill Assessment Using Unified Foundation Model}
%
\author{Satoshi Kondo}
%
\authorrunning{Satoshi Kondo}
%
\institute{Muroran Institute of Technology, Hokkaido, Japan\\
\email{kondo@muroran-it.ac.jp}}
%
\maketitle              
\begin{abstract}
Surgical skill assessment is paramount for ensuring patient safety and enhancing surgical outcomes. This study addresses the need for efficient and objective evaluation methods by introducing ZEAL (surgical skill assessment with \underline{Z}ero-shot surgical tool segmentation with a unifi\underline{E}d found\underline{A}tion mode\underline{L}). ZEAL uses segmentation masks of surgical instruments obtained through a unified foundation model for proficiency assessment. Through zero-shot inference with text prompts, ZEAL predicts segmentation masks, capturing essential features of both instruments and surroundings. Utilizing sparse convolutional neural networks and segmentation masks, ZEAL extracts feature vectors for foreground (instruments) and background. Long Short-Term Memory (LSTM) networks encode temporal dynamics, modeling sequential data and dependencies in surgical videos. Combining LSTM-encoded vectors, ZEAL produces a surgical skill score, offering an objective measure of proficiency. Comparative analysis with conventional methods using open datasets demonstrates ZEAL's superiority, affirming its potential in advancing surgical training and evaluation. This innovative approach to surgical skill assessment addresses challenges in traditional supervised learning techniques, paving the way for enhanced surgical care quality and patient outcomes.

\end{abstract}
\section{Introduction}

\noindent
Assessing surgical skill is crucial for ensuring patient safety and improving surgical outcomes. It helps identify areas for improvement and provides targeted feedback to surgeons, contributing to the continuous improvement of surgical care quality. According to a study, there is significant variation in technical performance among accredited surgeons, which is associated with clinical and pathological outcomes~\cite{curtis2020association}. The conventional method of evaluating surgical proficiency is through subjective observation and manual assessment by experienced surgeons. However, this approach is gradually being supplemented by automated techniques that utilize machine learning algorithms. This shift towards automated surgical skill assessment holds great promise for enhancing the efficiency, accuracy, and objectivity of surgical skill evaluation.

Numerous studies have demonstrated the significance of surgical instrument information in automated surgical skill assessment. Kim et al. proposed a method for assessing surgical skill in capsulectomy~\cite{kim2019objective}.  They utilized temporal convolutional neural networks~\cite{lea2017temporal} to estimate surgical skill (expert or novice) based on three features: tool tip velocities, tool tip positions, and flow fields in videos. The model's performance was evaluated by altering the combination of features. The study concluded that the model utilizing tip velocities was the most effective. Jin et al. introduced a method for automated evaluation of surgeon performance by tracking and analyzing tool movements in surgical videos~\cite{jin2018tool}. They utilized region-based convolutional neural networks, specifically Faster R-CNN~\cite{ren2015faster}. Liu et al. proposed a unified multi-pathway framework for automatic surgical skill assessment. The framework considers multiple aspects of surgical skill, including tool usage, intraoperative event patterns, and other skill proxies~\cite{liu2021towards}. The study reports the results of using single and multiple pathways.  When utilizing a single pathway, the proxy pathway yields superior outcomes compared to other path-ways in the clinical data. Combining multiple pathways generally improves performance.The optimal average performance is achieved when all pathways are utilized. Yanik et al. proposed a model that utilizes Mask R-CNN~\cite{he2017mask} to generate tool motion sequences from video frames. The sequences are then embedded using a denoising autoencoder (DAE) for the classifier to predict summative and formative performance~\cite{yanik2023video}.  Goldbraikh et al. proposed an algorithm that localizes tools and hands, identifies interactions between them, and computes motion metrics for surgical skill assessment~\cite{goldbraikh2022video}. Fathollahi et al. propose an approach for automatically assessing surgical skill from video feeds of surgical cases~\cite{fathollahi2022video}. The proposed pipeline tracks surgical tools to generate motion trajectories, which are then used to predict the technical skill level of the surgeon.

As previously stated, studies have emphasized the importance of surgical instrument information in automated surgical assessment. To obtain this information, a model for object detection or segmentation is required.  This process involves using supervised learning techniques, which require image datasets with object bounding boxes or object segmentation masks for training. However, it is important to note that preparing datasets with this level of annotation can be time-consuming and expensive.

This study introduces a novel approach called ZEAL (surgical skill assessment with \underline{Z}ero-shot surgical tool segmentation with a unifi\underline{E}d found\underline{A}tion mode\underline{L}) to evaluate surgical skills. The method leverages segmentation masks of surgical instruments to assess the proficiency of surgeons. ZEAL uses a unified foundation model to predict segmentation masks through zero-shot inference with text prompts. To extract meaningful information from the images, the proposed method extracts two feature vectors for each image: one for the foreground, which represents the surgical instruments, and another for the background. This is achieved by employing sparse convolutional neural networks using segmentation masks. These feature vectors capture the essential characteristics of both the surgical instruments and the surrounding environment. To capture the temporal dynamics of the surgical procedure, ZEAL encodes the time series of the foreground and background feature vectors using Long Short-Term Memory (LSTM) networks. LSTM networks are well-suited for modeling sequential data and can effectively capture the dependencies between consecutive frames in the surgical video. By combining the encoded feature vectors by the LSTM networks, ZEAL predicts a surgical skill score. This score serves as an objective measure of the surgeon's proficiency, providing valuable insights into their performance. The proposed method offers a promising approach to objectively assess surgical skills, paving the way for advancements in surgical training and evaluation.

Our contributions can be summarized as follows: Firstly, we introduce a unified foundation model that enables zero-shot inference for segmenting surgical instruments from surgical videos. Secondly, we use the segmentation results for surgical skill assessment. To the best of our knowledge, this is the first study to utilize foundation models for surgical skill assessment. Finally, we compare our proposed method with conventional methods using open datasets and confirm that it outperforms the conventional methods in one of the evaluation metrics.

\section{Proposed Method}

\begin{figure}[tb]
\centering
\includegraphics[scale=0.85]{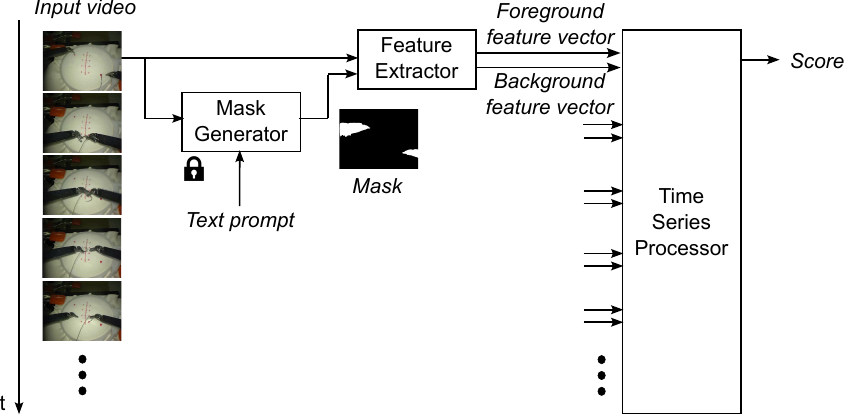}
\caption{Overview of ZEAL.}
\label{fig:outline}
\end{figure}

\noindent
Figure~\ref{fig:outline} outlines the proposed method called ZEAL for the estimation of the surgical skill score for a given surgical video. The method consists of three components: the mask generator, the feature extractor, and the time series processor. The mask generator and the feature extractor process each frame in the video separately, and the time series processor aggregates the features of all the frames in a video. The following paragraphs provides a detailed explanation of each component.

The mask generator estimates the regions of surgical instruments in an input image. We utilize the unified foundation model Grounded SAM~\cite{ren2024grounded} to achieve zero-shot inference for segmenting surgical instrument regions from images. Grounded SAM utilizes Grounding DINO~\cite{liu2023grounding} as an open-set object detector from text prompts to combine with the segment anything model (SAM)~\cite{kirillov2023segment}.  This integration allows the segmentation of regions based on any text input, regardless of its content. In our proposed method, text prompts, such as ``tool'', are provided to the grounded SAM in order to obtain segmentation masks for the corresponding regions. However, Grounding DINO sometimes fails to detect objects, which can have a negative impact on the performance of video analysis. To overcome this problem, we combine the bounding boxes of objects in the previous and current frames. The combined bounding boxes are provided to the SAM as the bounding box prompts.

The feature extractor is the second component. It takes an image and the corresponding mask generated by the mask generator as inputs. The mask is divided into square patches and resized to $w/p \times h/p$, where $w$ and $h$ are the width and height of the image, respectively, and $p$ is the size of the patch. The value of the mask for a patch is determined by the ratio of foreground pixels to all pixels in the patch, represented as $p \times p$. The values in the resized mask range from 0 to 1. The feature extractor employs CNN with sparse convolutions~\cite{woo2023convnext} to process an image and a mask as input. The feature extractor generates two feature vectors for each image. The first feature vector, which includes information about the foreground (i.e., surgical tools), is obtained from the image and the mask and will be referred to as the foreground feature vector. The second feature vector is generated by using the image and an inverted mask ($1-$mask) which includes information about the background, i.e., the image without surgical tools. This is referred to as the background feature vector.

Note that only the paths for the first image are shown in Fig.~\ref{fig:outline}. However, the mask generator and feature extractor process each image in the input video. The mask generator is fixed and not trained, while the feature extractor is shared across all images.

The time series processor takes as inputs the foreground feature vectors and the background feature vectors, both of which are time series. The feature vectors are encoded using bidirectional LSTM~\cite{graves2005framewise}. The feature vectors for the foreground and background are encoded independently. The output sequences from the bi-directional LSTM are then averaged using temporal average pooling for both foreground and background. The pooled features are concatenated and converted into a single float value, referred to as a score, using a linear layer.

\section{Experiments}

\subsection{Datasets and Experiment Settings}

\subsubsection{Datasets}
Experiments were conducted on the public JIGSAWS dataset~\cite{gao2014jhu}, which includes data on three elementary surgical tasks performed by surgeons on bench-top models. The tasks are Suturing (SU), Needle-Passing (NP), and Knot-Tying (KT). The JIGSAWS dataset includes 78 egocentric videos for the suturing task, 56 for needle passing, and 72 for knot tying, totaling 206 videos. On average, the videos have a duration of 88 seconds. The video data is annotated with a global rating score (GRS) that reflects the skill level. The GRS is calculated by summing up the scores of six elements, each rated on a Likert scale of 1 through 5, resulting in a range from 6 to 30. The modified objective structured assessments of technical skills (OSATS) approach~\cite{martin1997objective} is used for scoring. We considerd the GRS as the reference for surgical skill. The JIGSAWS dataset includes both kinematic and video data, but we only utilized the video data in our experiments.

\subsubsection{Evaluation Metrics}
We use two evaluation metrics to compare with previous work. The first metric is Spearman's rank correlation~\cite{parmar2017learning}, which is defined by the following equation:
\begin{equation}
\rho = 1 - \frac{6\sum_{i} d_{i}}{n(n^{2}-1)},
\end{equation}
where $d_{i}$ is the difference between the ranks of corresponding variables and $n$ is the number of samples. The second metric is the relative L2-distance (R-$\ell_{2}$)~\cite{yu2021group}. The equation for R-$\ell_{2}$ is defined as follows, given the highest and lowest scores for an action ($s_{max}$ and $s_{min}$):
\begin{equation}
\text{R-}\ell_{2} = \frac{1}{K} \sum_{k=1}^{K} \left( \frac{| s_{k} - \hat{s}_{k} |}{s_{max} - s_{min}} \right)^{2},
\end{equation}
where $s_{k}$ and $\hat{s}_{k}$ are used to represent the ground-truth and prediction scores for the $k$-th sample, respectively. It is important to note that R-$\ell_{2}$ differs from Spearman's correlation. While Spearman's correlation focuses more on the ranks of the predicted scores, R-$\ell_{2}$ focuses on the numerical values.

\subsubsection{Experiment Settings}
The method was evaluated on each task using a 4-fold cross-validation procedure~\cite{liu2021towards}. The videos used in each fold were identical to those used in~\cite{musdl_github,tang2020uncertainty}. Validation videos were selected from the training dataset, and the ratio of training, validation, and test data was 2:1:1 for each trial. We uniformly sampled 160 frames for each video, following the procedure outlined in \cite{tang2020uncertainty}.

The text prompt provided to the mask generator was ``metallic tool''. Several text prompts were tested, including ``tool'' and ``surgical tool'', among others. Ultimately, ``metallic tool'' was chosen based on objective evaluation of segmentation performance.

The metrics for the test data were calculated using the best model, which was determined by the lowest loss value for validation data during the training. The learning rate was adjusted in each epoch using cosine annealing with warmup. The maximum learning rate after the warmup period, which was a hyper-parameter determined through grid search, was $3 \times 10^{-5}$. 

The sparse convolutional network used was ConvNeXt-N with 15.6M parameters, and the bidirectional LSTM had one layer. The patch size $p$ was 32. The training data augmentations include shift, scale, rotation, color jitter, Gaussian blur, and Gaussian noise. All images are resized to 224$\times$224 pixels and normalized.
During the training phase, the experimental conditions consisted of 200 epochs, with a warm-up period of 20 epochs, and the utilization of the AdamW optimizer~\cite{loshchilov2017decoupled}.

The method was implemented using PyTorch v2.1.2~\cite{paszke2019pytorch} and Lightning v2.1.3 on an Nvidia RTX4090 GPU with 24 GB memory.

Our proposed method was compared to several existing methods~\cite{bai2022action,li2022surgical,tang2020uncertainty,yu2021group,zhang2023auto,zhou2022uncertainty}. The experiments in the conventional methods were conducted using the same 4-fold cross-validation procedure and evaluation metrics in our settings. However, it is important to note that the authors do not provide instructions on how to select validation videos from training videos or which model to use in order to obtain metrics for test videos.
As part of an ablation study, we evaluated our proposed method by replacing the bidirectional LSTM layer in the time series processor with a temporal pooling layer and a 3-layer MLP.

\subsection{Results}
\begin{figure}[tb]
  \centering
  \begin{minipage}{0.9\linewidth}
  \centering
  \includegraphics[scale=0.2, bb=0 0 640 480]{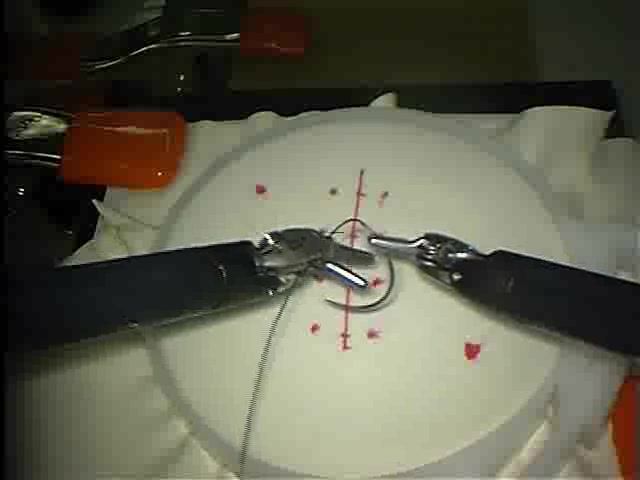}
  \includegraphics[scale=0.2, bb=0 0 640 480]{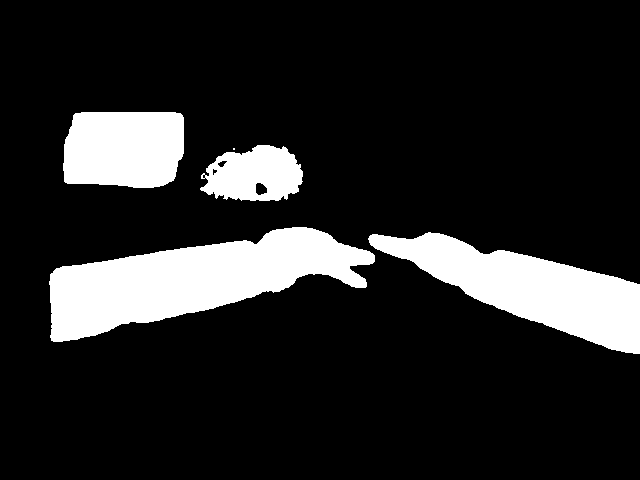}
    
  (a)
  \end{minipage}

  \vspace{.5\baselineskip}

  \centering
  \begin{minipage}{0.9\linewidth}
  \centering
  \includegraphics[scale=0.2, bb=0 0 640 480]{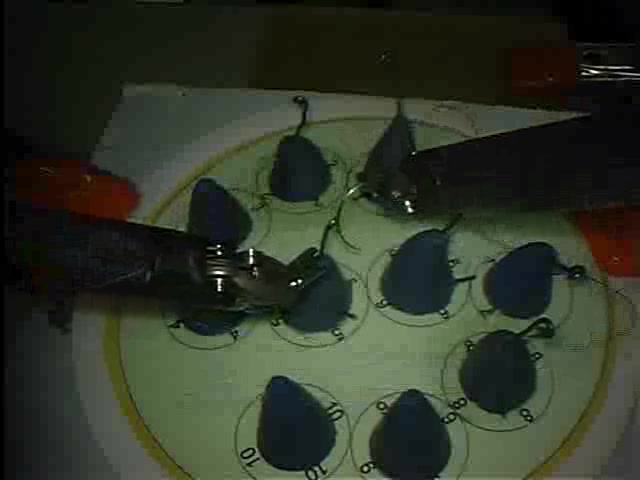}
  \includegraphics[scale=0.2, bb=0 0 640 480]{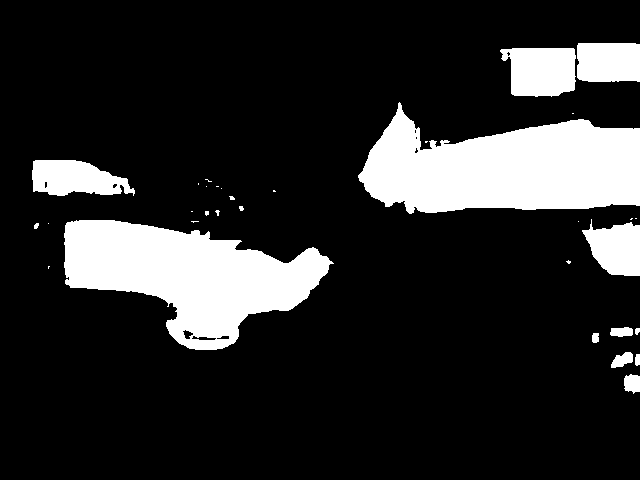}
    
  (b)
  \end{minipage}

  \vspace{.5\baselineskip}

  \centering
  \begin{minipage}{0.9\linewidth}
  \centering
  \includegraphics[scale=0.2, bb=0 0 640 480]{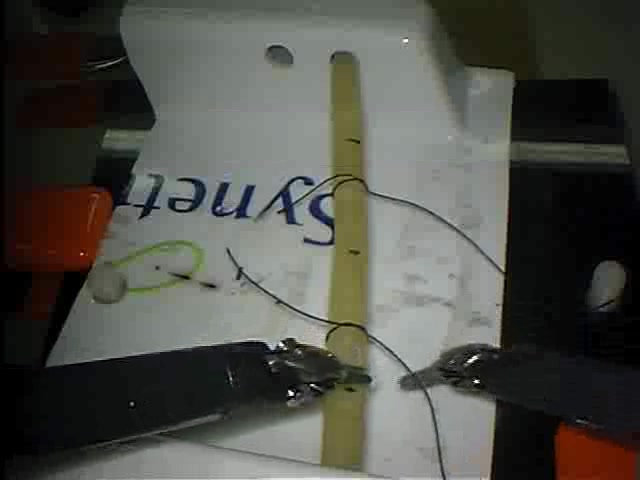}
  \includegraphics[scale=0.2, bb=0 0 640 480]{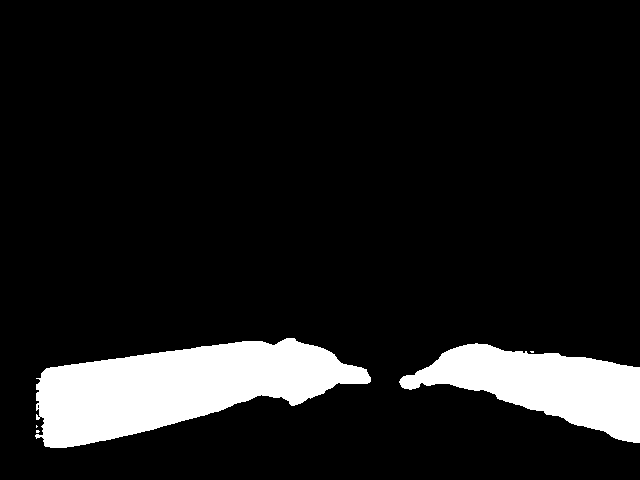}
  
  (c)
  \end{minipage}

  \caption{Segmentation results by the unified foundation model with the text prompt ``metallic tool''. (a) Surturing. (b) Needle Passing. (c) Knot Tying.}
  \label{fig:sam}
\end{figure}

Figure~\ref{fig:sam} shows some examples of the segmentation masks obtained by the unified foundation model with the text prompt ``metallic tool''. As can be seen in Figure 2, surgical tools are well segmented, but the boundaries of the masks are not clear. And other objects are also segmented. Exact segmentation masks are not necessary in our proposed method as convolutional neural networks can extract important features from the masked images, including foreground and background.

Table~\ref{tab:1} shows comparisons of performance with existing methods in the Spearman's rank correlation metric. Table~\ref{tab:2} shows the comparisons of performance with existing methods in the R-$\ell_{2}$ metric.

Table~\ref{tab:1} shows that the proposed method is not competitive with the existing methods, according to the evaluation using Spearman's rank correlation metric. On the other hand, Table~\ref{tab:2} shows that the proposed method outperforms the existing methods, including the state-of-the-art (SOTA), in the evaluation using the R-$\ell_{2}$ metric. Specifically, it outperforms UDAQA (the current SOTA that does not use exemplars) by approximately 1.2 (26.9 \% improvement) and TPT (the current SOTA that uses exemplars) by approximately 0.5 (13.7 \% improvement), in the R-$\ell_{2} \times 100$ metric respectively. 

Regarding the comparison of time series processors, specifically bidirectional LSTM and temporal pooling with MLP, bi-directional LSTM exhibits slightly better performance than temporal pooling with MLP for both metrics. This can be observed in Tables~\ref{tab:1} and \ref{tab:2}.

\begin{table}[t]
\caption{Comparisons of performance with existing methods (Spearman's rank correlation).
}
\label{tab:1}

\begin{center}
\begin{tabular}{c|c|c|c|c|c} \bhline{1.2pt}
Method & Exemplar & SU & NP & KT & Average $\uparrow$  \\\bhline{1.2pt}
USDL \cite{tang2020uncertainty} & No & 0.64 & 0.63 & 0.61 & 0.63 \\\hline
MUSDL \cite{tang2020uncertainty} & No & 0.71 & 0.69 & 0.71 & 0.70 \\\hline
I3D + MLP \cite{yu2021group} & No & 0.61 & 0.68 & 0.66 & 0.65 \\\hline
CoRe + GART \cite{yu2021group} & Yes & 0.84 & 0.86 & 0.86 & 0.85 \\\hline
ViSA \cite{li2022surgical} & No & 0.84 & 0.86 & 0.79 & 0.83 \\\hline
DAE-CoRe \cite{zhang2023auto} & No & 0.86 & 0.86 & 0.87 & 0.86 \\\hline
UDAQA \cite{zhou2022uncertainty} & No & 0.87 & 0.93 & 0.86 & {\bf 0.89} \\\hline
TPT \cite{bai2022action} & Yes & 0.88 & 0.88 & 0.91 & {\bf 0.89} \\\bhline{1.2pt}
Ours (ZEAL (Temporal Pool)) & No &  0.61 &  0.64 & 0.47 &  0.57 \\\hline
Ours (ZEAL (BiLSTM)) & No & 0.62 &  0.67 &  0.54 &  0.61 \\\bhline{1.2pt}
\end{tabular}
\end{center}
\end{table}

\begin{table}[t]
\caption{Comparisons of performance with existing methods (R-$\ell_{2} \times 100$).
 }
\label{tab:2}

\begin{center}
\begin{tabular}{c|c|c|c|c|c} \bhline{1.2pt}
Method & Exemplar & SU & NP & KT & Average $\downarrow$ \\\bhline{1.2pt}
I3D + MLP \cite{yu2021group} & No & 4.795 & 11.225 & 6.120 & 7.373 \\\hline
CoRe + GART \cite{yu2021group} & Yes & 5.055 & 5.688 & 2.927 & 4.556 \\\hline
UDAQA \cite{zhou2022uncertainty} & No & 3.444 & 4.076 & 5.469 & 4.330 \\\hline
TPT \cite{bai2022action} & Yes & 2.722 & 5.259 & 3.022 & 3.668 \\\bhline{1.2pt}
Ours (ZEAL (Temporal Pool)) & No &  3.325 &  2.550 &  3.650 &  3.175 \\\hline
Ours (ZEAL (BiLSTM)) & No & 3.150 & 2.675 & 3.675 &  {\bf 3.167} \\\bhline{1.2pt}
\end{tabular}
\end{center}
\end{table}

\section{Conclusions}

We proposed a novel approach called ZEAL for evaluating surgical skills. ZEAL uses a unified foundation model to predict segmentation masks through zero-shot inference with text prompts. The proposed method extracts two feature vectors for each image, one for the foreground and another for the background, using sparse convolutional neural networks. ZEAL encodes the time series of the foreground and background feature vectors using LSTM networks. 

The experiments conducted on the JIGSAW dataset indicated that the proposed method was not as competitive as existing methods, as evaluated by Spearman's rank correlation metric. However, according to the evaluation by the R-$\ell_{2}$ metric, the proposed method outperformed existing methods, including the state-of-the-art (SOTA).

Future directions for improvement of the proposed method include enhancing performance the use of alternative loss functions, such as Spearman's rank correlation loss. Another direction is to assess the effectiveness of the proposed method using clinical datasets.

\newpage
%
%
\bibliographystyle{splncs04}
\bibliography{surgical_skill}
%
%
%
%
%

\end{document}